\documentclass[11pt,a4paper]{article}
\usepackage[hyperref]{acl2020}
\usepackage{times}
\usepackage{latexsym}

\usepackage{microtype}
\usepackage{amsmath}
\usepackage{amssymb}
\usepackage{commath}
\usepackage{csquotes}
\usepackage{mathtools}
\usepackage{esvect}
\usepackage{amsfonts}
\usepackage{color}
\newcommand{\inner}[2]{\langle #1,#2 \rangle}

\aclfinalcopy 
\def\aclpaperid{2141} 

\title{Stolen Probability: A Structural Weakness of Neural Language Models}

\author{David Demeter \\
  Northwestern University \\
  \texttt{ddemeter@u.northwestern.edu}
  \\\And
  Gregory Kimmel \\
  H. Lee Moffitt Cancer Center \\
  \texttt{gregory.kimmel@moffitt.org} \\
  \\\AND
  Doug Downey \\
  Allen Institute for AI \\
  \texttt{dougd@allenai.org} \\}

\date{}

\begin{document}
\maketitle

\begin{abstract}
Neural Network Language Models (NNLMs) generate probability distributions by applying a softmax function to a distance metric formed by taking the dot product of a prediction vector with all word vectors in a high-dimensional embedding space.  The dot-product distance metric forms part of the inductive bias of NNLMs.  Although NNLMs optimize well with this inductive bias, we show that this results in a sub-optimal ordering of the embedding space that structurally impoverishes some words at the expense of others when assigning probability.  We present numerical, theoretical and empirical analyses showing that words on the interior of the convex hull in the embedding space have their probability bounded by the probabilities of the words on the hull.

\end{abstract}

\section{Introduction}

Neural Network Language Models (NNLMs) have evolved rapidly over the years from simple feed forward nets \cite{Bengio03aneural} to include recurrent connections \cite{Mikolov2010RecurrentNN} and LSTM cells \cite{Zaremba2014RecurrentNN}, and most recently transformer architectures \cite{Dai2019TransformerXLAL, Radford2019LanguageMA}.  This  has enabled ever-increasing performance on benchmark data sets. However, one thing has remained relatively constant: the softmax of a dot product as the output layer.

NNLMs generate probability distributions by applying a softmax function to a distance metric formed by taking the dot product of a prediction vector with all word vectors in a high-dimensional embedding space.  We show that the dot product distance metric introduces a limitation that bounds the expressiveness of NNLMs, enabling some words to \enquote{steal} probability from other words simply due to their relative placement in the embedding space.  We call this limitation the {\em stolen probability effect}.  While the net impact of this limitation is small in terms of the perplexity measure on which NNLMs are evaluated, we show that the limitation results in significant errors in certain cases.

As an example, consider a high probability word sequence like \enquote{the United States of America} that ends with a relatively infrequent word such as \enquote{America}. Infrequent words are often associated with smaller embedding norms, and may end up inside the convex hull of the embedding space.  As we show, in such a case it is impossible for the NNLM to assign a high probability to the infrequent word that completes the high-probability sequence.

Numerical, theoretical and empirical analyses are presented to establish that the stolen probability effect exists.  Experiments with n-gram models, which lack this limitation, are performed to quantify the impact of the effect. 

\section{Background}

In a NNLM, words $w_i$ are represented as vectors $x_i$ in a high-dimensional embedding space. Some combination of these vectors $x_c = \{x_i\}_{i \in c}$ are used to represent the preceding context $c$, which are fed into a a neural unit as features to generate a prediction vector $h_t$.  NNLMs generate a probability distribution over a vocabulary of words $w_i$ to predict the next word in a sequence $w_t$ using a model of the form:

\begin{align}\label{eq:nnlm}
P(w_t|c) = \sigma (f(x_c,\theta_{NNLM}))
\end{align}

\noindent where $\sigma$ is the softmax function, $f$ is a neural unit that generates the prediction vector $h_t$, and $\theta_{NNLM}$ are the parameters of the neural unit.

A dot product between the prediction vector $h_t$ and all word vectors $x_i$ is taken to calculate a set of distances, which are then used to form {\em logits}:

\begin{align}\label{eq:logit}
z_{it} = x_i \cdot h_t^T + b_i
\end{align}

\noindent where $b_i$ is a word-specific bias term.  Logits are used with the softmax function to generate a probability distribution over the vocabulary $V$ such that:

\begin{align}\label{eq:softmax}
P(w_t=w_i|c)=\frac{e^{z_{it}}}{\sum_{V}e^{z_{vt}}}
\end{align}

\noindent We refer to this calculation of logits and transformation into a probability distribution as the {\em dot-product softmax}.

\section{Problem Definition}

NNLMs learn very different embeddings for different words.  In this section we show that this can make it impossible for words with certain embeddings to {\em ever} be assigned high probability in {\em any} context.  We start with a brief examination of the link between embedding norm and probability, which motivates our analysis of the stolen probability effect in terms of a word's position in the embedding space relative to the convex hull of the embedding space.

\subsection{Embedding Space Analysis}

The dot product used in Eq. \ref{eq:logit} can be written in polar coordinates as:

\vspace{-0.1in}
\begin{align}\label{eq:polar}
z_{it} = \norm{x_i} \norm{h_t} cos(\theta_i) + b_i
\end{align}

\noindent where $\theta_i$ is the angle between $x_i$ and $h_t$.  The dot-product softmax allocates probability to word $w_i$ in proportion to $z_{it}$'s value relative to the value of other logits (see Eq. \ref{eq:softmax}).  Setting aside the bias term $b_i$ for the moment (which is shown empirically to be irrelevant to our analysis in Section \ref{sec:results}), this means that word $A$ with a larger norm than word $B$ will be assigned higher probability when the angles $\theta_A$ and $\theta_B$ are the same.

More generally, the relationship between embedding norms and the angles formed with prediction points $h_t$ can be expressed as:

\begin{align}\label{eq:example}
\frac{\norm{x_A}}{\norm{x_B}} > \frac{cos(\theta_B)}{cos(\theta_A)}
\end{align}

\noindent when word $A$ has a higher probability than word $B$.  Empirical results (not presented) confirm that NNLMs organize the embedding space such that word vector norms are widely distributed, while their angular displacements relative to a reference vector fall into a narrow range.  This suggests that the norm terms in Eq. \ref{eq:polar} dominate the calculation of logits, and thereby probability.

\subsection{Theoretical Analysis}

While an analysis of how embedding norms impact the assignment of probability is informative, the stolen probability effect is best analyzed in terms of a word's position in the embedding space relative to the convex hull of the embedding space.  A convex hull is the smallest set of points forming a convex polygon that contains all other points in a Euclidean space.

{\bf Theorem 1.} {\em Let $C$ be the convex hull of the embeddings $\{x_i\}$ of a vocabulary $V$.  If an embedding $x_i$ for word $w_i \in V$ is interior to $C$, then the maximum probability $P(w_i)$ assigned to $w_i$ using a dot-product softmax is bounded by the probability assigned to at least one word $w_i$ whose embedding is on the convex hull.} (see Appendix A for proof).

\begin{figure}
\includegraphics[scale=0.43]{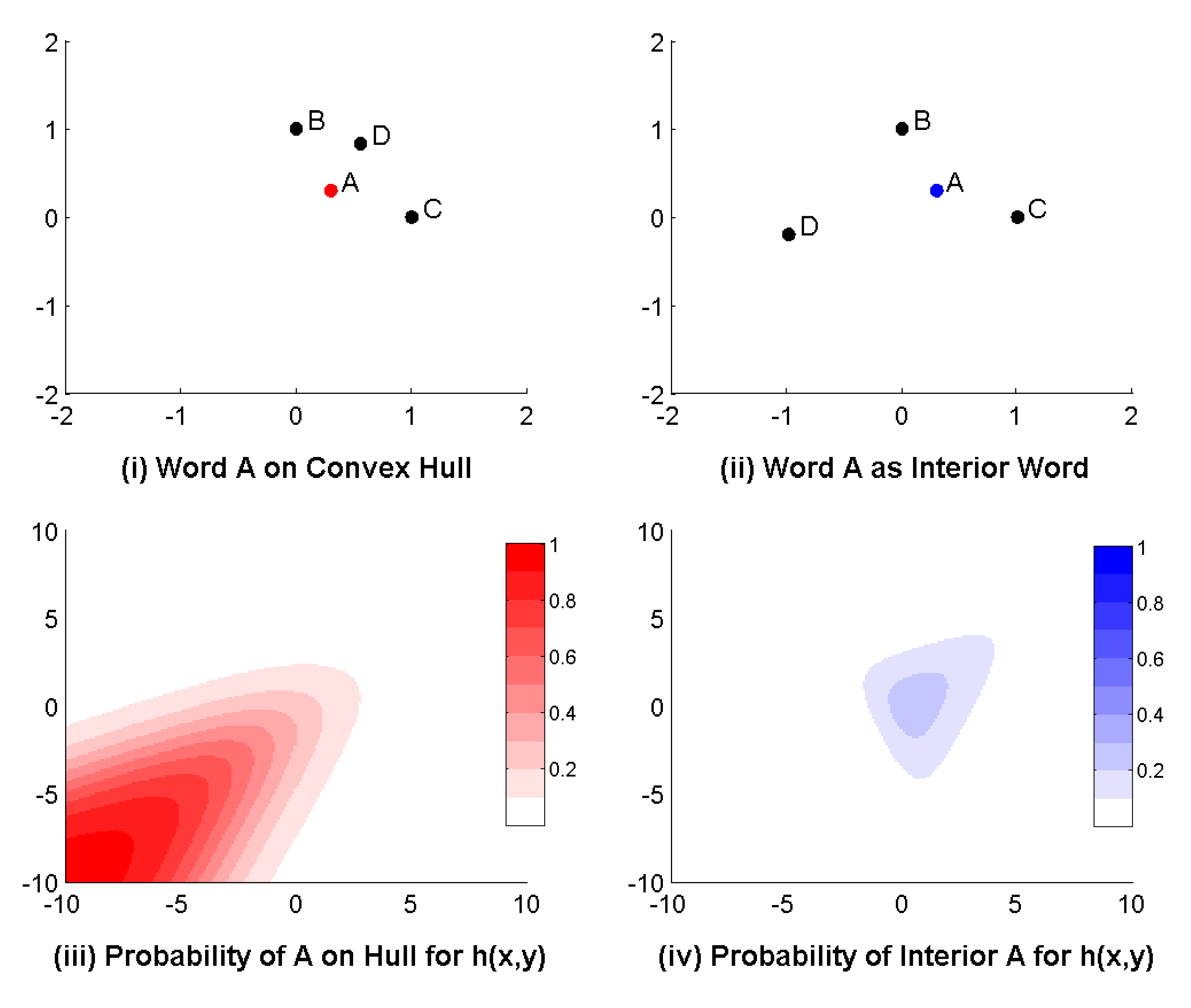}
\centering
\caption{\textbf{Numerical Illustration of the Stolen Probability Effect.} {Panels (i) and (ii) show the embedding of four words in a 2D embedding space.  Word $A$ is on the convex hull in panel (i), and interior to the convex hull in panel (ii).  Panels (iii) and (iv) show the probability that would be assigned by the dot-product softmax to $A$ for a range of prediction points $h_t$ in the $x,y$ plane.  When word $A$ is on the convex hull, it can achieve nearly 100\% probability for an $h_t$ prediction point in the far lower-left quadrant (see panel (iii)).  When word $A$ is interior to the convex hull, its maximum probability is bounded by any word on the convex hull (see panel (iv)).}}
\label{fig:f_matlab}
\end{figure}

\subsection{Numerical Analysis}

The stolen probability effect can be illustrated numerically in a 2D Euclidean space (see Figure \ref{fig:f_matlab}).  We show two configurations of an embedding space, one where target word $A$ is on the convex hull (Panel i) and another where $A$ is on the interior (Panel ii).  Under both configurations, a NNLM trained to the maximum likelihood objective would seek to assign probability such that $P(A) = 1.0$.
 
For the first configuration, this is achievable for an $h_t$ in the far lower-left quadrant (Panel iii).  However, when $A$ is in the interior, there is no $h_t$ that exists where the dot-product softmax can assign a probability approaching $1.0$ (Panel iv).  A similar illustration in 3D is presented in Appendix B.

\section{Experiments}

In this section we provide empirical evidence showing that words interior to the convex hull are probability-impoverished due to the stolen probability effect and analyze the impact of this phenomenon on different models.

\subsection{Methods}
 
 We perform our evaluations using the AWD-LSTM \cite{Merity2017RegularizingAO} and the Mixture of Softmaxes (MoS) \cite{Yang2017BreakingTS} language models.  Both models are trained on the Wikitext-2 corpus  \cite{Merity2016PointerSM} using default hyper-parameters, except for dimensionality which is set to $d = \{50,100,200\}$.  The AWD-LSTM model is trained for 500 epochs and the MoS model is trained for 200 epochs, resulting in perplexities as shown in Table \ref{tab:ppl}.  

 The Quickhull algorithm \cite{Barber1996TheQA} is among the most popular algorithms used to detect the convex hull in Euclidean space.  However, we found it to be intractably slow for embedding spaces above ten dimensions, and therefore resorted to approximate methods.  We relied upon an identity derivable from the properties of a convex hull which states that {\em a point $p \in \mathbb{R}^d$ is vertex of the convex hull of $\{x_i\}$ if there exists a vector $h_t \in \mathbb{R}^d$ such that for all $x_i$:}

\begin{align}\label{eq:detect}
\langle h_t,x_i-p \rangle < 0.
\end{align}

\noindent where $\langle \cdot \rangle$ is the dot-product.

Searching for directions $h_t$ which satisfy Eq \ref{eq:detect} is not computationally feasible.  Instead, we rely upon a high-precision, low-recall approximate method to eliminate potential directions for $h_t$ which do not satisfy Eq. \ref{eq:detect}.  We call this method our {\em detection algorithm}.  If the set of remaining directions is not empty, then $p$ is classified as a vertex, otherwise $p$ is classified as an interior point.  

The detection algorithm is anchored by the insight that all vectors parallel to the difference vector $\vec{x_i} - \vec{p}$ do not satisfy Eq. \ref{eq:detect}.  It is also true that all directions in the range ($\phi + \omega,\phi - \omega$) will not satisfy Eq. \ref{eq:detect}, where $\phi$ is the direction of the difference vector and $\omega$ is some increment less than $\pi/2$.  The detection algorithm was validated in lower dimensional spaces where an exact convex hull could be computed ({\em e,g.} up to $d = 10$).  It consistently classified interior points with precision approaching 100\% and recall of 68\% when evaluated on the first $10$ dimensions of the MoS model with $d = 100$.

\begin{table}
\small
\begin{tabular}{cccccc}
\hline
\rule{0pt}{3ex} & & {\bf Train} & {\bf Test} & {\bf $\omega$} & {\bf Interior}\\
{\bf Model} & {\bf {\em d}} & {\bf PPL} & {\bf PPL} & {\bf (radians)} & {\bf Points} \\
\hline
\rule{0pt}{3ex}AWD & 50 & 140.6 & 141.8 & {\small $50\pi / 128$} & 6,155\\
AWD & 100 & 73.3 & 97.8 & {\small $55\pi / 128$} & 5,205\\
AWD & 200 & 44.9 & 81.6 & {\small $58\pi / 128$} & 2,064\\
\rule{0pt}{3ex}MoS & 50 & 51.7 & 76.8 & {\small $53\pi / 128$} & 4,631\\
Mos & 100 & 34.8 & 67.4 &  {\small $57\pi / 128$} & 4,371\\
MoS & 200 & 25.5 & 64.2 & {\small $59\pi / 128$} & 2,009\\

\hline
\end{tabular}
\caption{\textbf{Perplexities and Detection Results.} Each model was trained using default hyper-parameters except for dimensions $d$ as shown and number of training epochs.  The AWD-LSTM models we trained for 500 epochs and the MoS models were trained for 200 epochs.  Each ordinal plane of an {\em n-Sphere} in the embedding space was discretized into arcs of $2\pi / 256$.  The angle $\phi$ of the difference vector $x_i-p$ formed each word type embedded at $p$ is mapped to one of these arcs.  Directions on the interval $(\phi \pm \omega)$ are eliminated from consideration per Eq \ref{eq:detect}, and words for which all directions have been eliminated as classified as interior. The increment $\omega$ was set to the lowest value that would classify at least 1,000 words as interior.}
\label{tab:ppl}
\end{table}

\subsection{Results}\label{sec:results}

\begin{figure}
\includegraphics[scale=0.30]{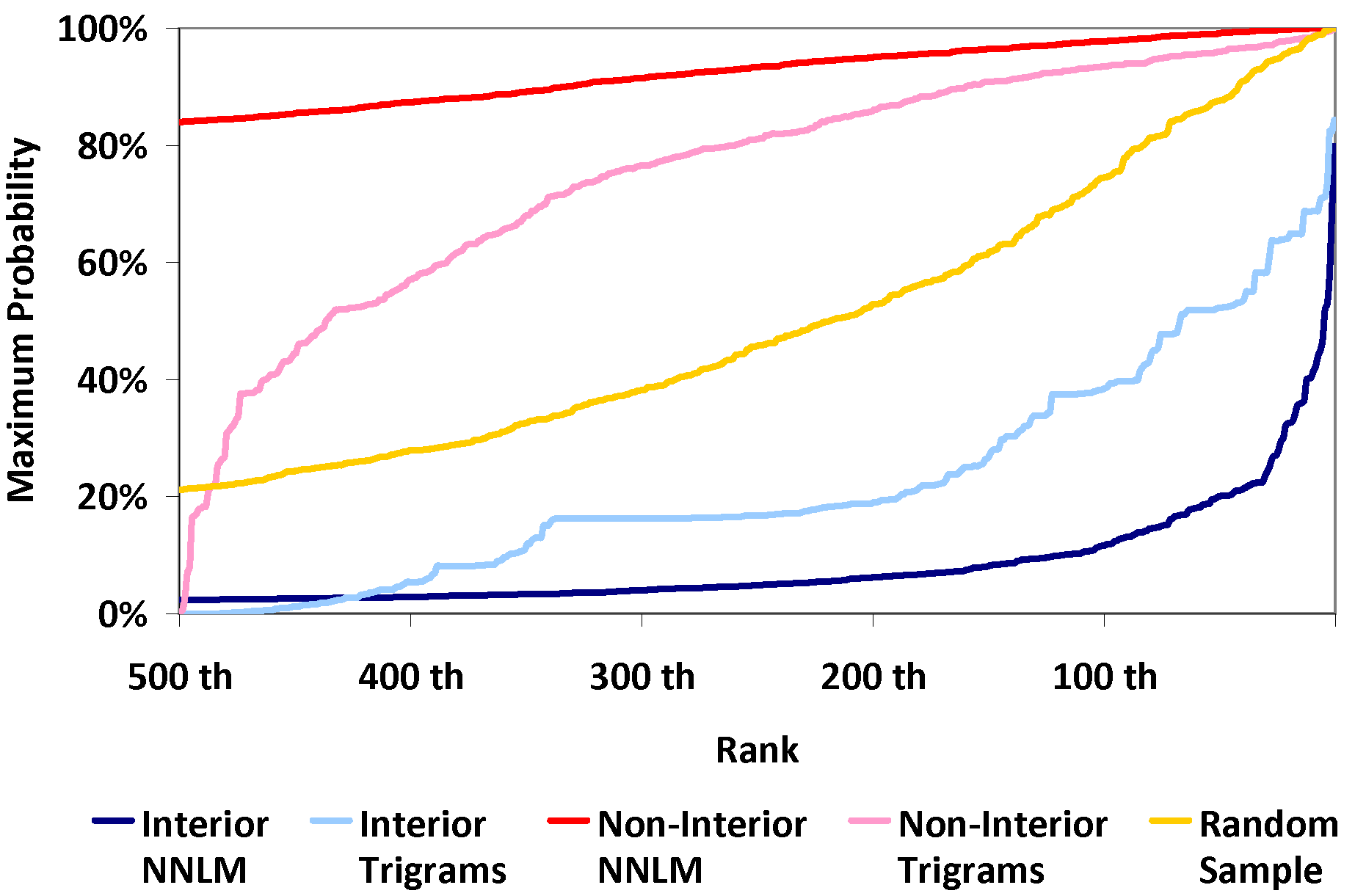}
\centering
\caption{\textbf{Maximum Probability of Top 500 Interior and Non-Interior Words.} {The MoS model with $d=100$ struggles to assign high probability to interior words, while trigrams were able to capture more accurate statistics.  This behavior is absent for non-interior words.}}
\label{fig:f_main}
\end{figure}

Applying the detection algorithm to our models yields word types being classified into distinct {\em interior} and {\em non-interior} sets (see Table \ref{tab:ppl}).  We ranked the top 500 words of each set by the maximum probability they achieved on the training corpora\footnote{We present our results on the training set because here, our goal is to characterize the expressiveness of the models rather than their ability to generalize.}, and plot these values in Figure \ref{fig:f_main}, showing a clear distinction between interior and non-interior sets.  The maximum trigram probabilities \cite{Stolcke2002SRILMA} smoothed with KN3 for the same top 500 words in each set (separately sorted) are also shown.  The difference between NNLM and trigram curves for interior words shows that models like n-grams, which do not utilize a dot-product softmax, are {\em not} subject to the stolen probability effect and can assign higher probabilities.  A {\em random} set of words equal in size to the interior set was also constructed by uniform sampling, and ranked on the top 500 words.  A comparison between the random and interior sets provides evidence that our detection algorithm is effective at separating the interior and non-interior sets, and is not simply performing random sampling. 

Our results can be more compactly presented by considering the average probability mass assigned to the top 500 words for each set (see Table \ref{tab:table}).  The impact of the stolen probability effect for each model can quantified as the difference between the interior set and each of the three reference sets (non-interior, random, and trigram) in the table.  The interior average maximum probability is generally much smaller than those of the reference sets.

\begin{table}
\small
\begin{tabular}{ccrrrr}
\hline
 &  & {\bf Non-} & & {\bf Tri-} & \\
{\bf Model} & {\bf $d$} & {\bf Interior} & {\bf Rand} & {\bf gram} & {\bf Interior}\\
\hline
\rule{0pt}{3ex}AWD & 50 & 44.3 & 8.1 & 20.7 & 0.004\\
AWD &  100 & 89.2 & 31.3 & 15.6 & 0.018\\
AWD &  200 & 99.0 & 43.3 & 12.5 & 0.113 \\
\rule{0pt}{3ex}MoS &  50 & 76.5 & 22.9 & 16.8 & 0.4\\
MoS &  100 & 92.9 & 50.5 & 22.6 & 8.6\\
MoS &  200 & 97.3 & 51.4 & 30.9 & 40.0\\
\hline
\end{tabular}
\caption{\textbf{Average Maximum Probability for Top 500 Words.}  The average probability mass for each word set (expressed as percents) is calculated by averaging the maximum probability on the training corpora achieved for the top 500 words of each set.}
\label{tab:table}
\end{table}

Another way to quantify the impact of the stolen probability effect is to overcome the bound on the interior set by constructing an ensemble with trigram statistics.  We constructed a targeted ensemble of the MoS model with $d = 100$ and a trigram model---unlike a standard ensemble, the trigram model is only used in contexts that are likely to indicate an interior word: specifically, those that precede at least one interior word in the training set.  Otherwise, we default to the NNLM probability.  When we ensemble, we assign weights of 0.8 to the NNLM, 0.2 to the trigram (selected using the training set).  Overall, the targeted ensemble improved training perplexity from 34.8 to 33.6, and test perplexity from 67.4 to 67.0.  The improvements on the interior words themselves were much larger: training perplexities for interior words improved from 700.0 to 157.2, and test improved from 645.6 to 306.7.  Improvement on the interior words is not unexpected given the differences observed in Figure \ref{fig:f_main}.  The overall perplexity differences, while small in magnitude, suggest that ensembling with a model that lacks the stolen probability limitation may provide some boost to a NNLM.

Returning to the question of bias terms, we find empirically that bias terms are relatively small, averaging $-0.13$ and $0.02$ for the interior and non-interior sets of the MoS model with $d = 100$, respectively.  We note that the bias terms are word-specific and can only adjust the stolen probability effect by a constant factor.  That is, it does not change the fact that words in the interior set are probability-bounded.  All of our empirical results are calculated on a model with a bias term, demonstrating that the stolen probability effect persists with bias terms.

\subsection{Analysis}

Attributes of the stolen probability effect analyzed in this work are distinct from the {\em softmax bottleneck} \cite{Yang2017BreakingTS}.  The softmax bottleneck argues that language modeling can be formulated as a factorization problem, and that the resulting model's expressiveness in limited by the rank of the word embedding matrix. While we also argue that the expressiveness of a NNLM is limited for structural reasons, the stolen probability effect that we study is best understood as a property of the {\em arrangement} of the embeddings in space, rather than the dimensionality of the space.

Our numerical and theoretical analyses presented do not rely upon any particular number of dimensions, and our experiments show that the stolen probability effect holds over a range of dimensions.  However, there is a steady increase of average probability mass assigned to the interior set as model dimensionality increases, suggesting that there are limits to the stolen probability effect.  This is not unexpected.  As the capacity of the embedding space increases with additional dimensions, the model has additional degrees of freedom in organizing the embedding space.  The vocabulary of the Wikitext-2 corpus is small compared to other more recent corpora.  We believe that larger vocabularies will offset (at least partially) the additional degrees of freedom associated with higher dimensional embedding spaces.  We leave the exploration of this question as future research.

We acknowledge that our results can also be impacted by the approximate nature of our detection algorithm.  Without the ability to precisely detect detect the convex hull for any of our embedding spaces, we can not make precise claims about its performance.  The difference between average probability mass assigned to random and interior sets across all models evaluated suggests that the detection algorithm succeeds at identifying words with substantially lower maximum probabilities than a random selection of words.

\begin{figure}
\includegraphics[scale=0.30]{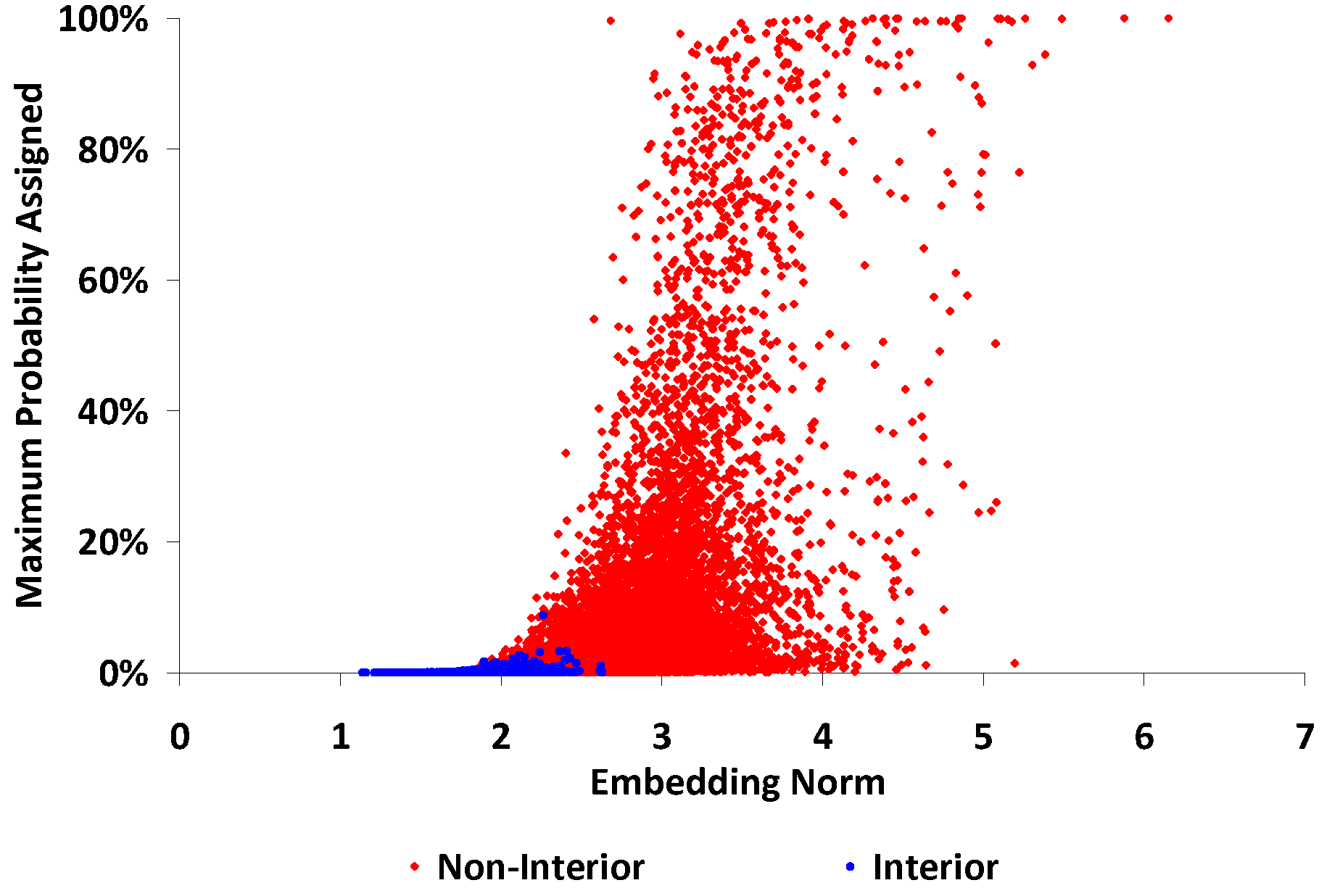}
\centering
\caption{\textbf{Maximum Probability vs. Embedding Norm.} {Examining maximum word probability as a function of embedding norm for the MoS model with $d = 100$ shows that interior words are associated with smaller embedding norms and lower maximum probabilities.  However, many non-interior words with comparably small norms have substantially higher maximum probabilities.}}
\label{fig:f_scatter}
\end{figure}

In Section 3.1 we motivated our analysis of the stolen probability effect by examining the impact of embeddings norms on probability assignment.  One natural question is to ask is {\em \enquote{Does our detection algorithm simply classify embeddings with small norms as interior points?}}  Our results suggest that this is not the case.  The scatter plot of embedding norm versus maximum probability (see Figure \ref{fig:f_scatter}) shows that words classified as interior points frequently have lower norms.  This is expected, since points interior to the convex hull are by definition not located in extreme regions of the embedding space.  The embedding norms for words in the interior set range between 1.4 and 2.6 for the MoS model with $d = 100$.  Average maximum probabilities for words in this range are $1.4\%$ and $4.1\%$ for interior and non-interior sets of the MoS model with $d = 100$, respectively, providing  evidence that the detection algorithm is not merely identifying word with small embedding norms.

Lastly, we note that the interior sets of the AWD-LSTM models are particularly probability impoverished relative to the more powerful MoS models.  We speculate that the perplexity improvements of the MoS model may be due in part to mitigating the stolen probability effect.  Exploration of the stolen probability effect in more powerful NNLM architectures using dot-product softmax output layers is another item of future research.

\section{Related Work}

Other work has explored alternative softmax configurations, including a mixture of softmaxes, adaptive softmax and a Taylor Series softmax \cite{Yang2017BreakingTS, Grave2016EfficientSA, Brbisson2015AnEO}.  There is also a body of work that analyzes the properties of embedding spaces \cite{Burdick2018FactorsIT, Mimno2017TheSG}.  We do not seek to modify the softmax.  Instead we present an analysis of how the structural bounds of an NNLM limit its expressiveness.

\section{Conclusion}

We present numerical, theoretical and empirical analyses showing that the dot-product softmax limits a NNLM's expressiveness for words on the interior of a convex hull of the embedding space.  This is structural weakness of NNLMs with dot-product softmax output layers, which we call the {\em stolen probability effect}.  Our experiments show that the effect is relatively common in smaller neural language models. Alternative architectures that can overcome the stolen probability effect are an item of future work.

\section*{Acknowledgments}

This work was supported in part by NSF Grant IIS-1351029.  We thank the anonymous reviewers and Northwestern's Theoretical Computer Science group for their insightful comments and guidance.

\bibliography{acl2020}

\begin{thebibliography}{14}
\expandafter\ifx\csname natexlab\endcsname\relax\def\natexlab#1{#1}\fi

\bibitem[{Barber et~al.(1996)Barber, Dobkin, and Huhdanpaa}]{Barber1996TheQA}
C.~Bradford Barber, David~P. Dobkin, and Hannu Huhdanpaa. 1996.
\newblock The quickhull algorithm for convex hulls.
\newblock \emph{ACM Trans. Math. Softw.}, 22:469--483.

\bibitem[{Bengio et~al.(2003)Bengio, Ducharme, Vincent, and
  Jauvin}]{Bengio03aneural}
Yoshua Bengio, Réjean Ducharme, Pascal Vincent, and Christian Jauvin. 2003.
\newblock A neural probabilistic language model.
\newblock \emph{JOURNAL OF MACHINE LEARNING RESEARCH}, 3:1137--1155.

\bibitem[{de~Br{\'e}bisson and Vincent(2015)}]{Brbisson2015AnEO}
Alexandre de~Br{\'e}bisson and Pascal Vincent. 2015.
\newblock An exploration of softmax alternatives belonging to the spherical
  loss family.
\newblock \emph{CoRR}, abs/1511.05042.

\bibitem[{Burdick et~al.(2018)Burdick, Kummerfeld, and
  Mihalcea}]{Burdick2018FactorsIT}
Laura Burdick, Jonathan~K. Kummerfeld, and Rada Mihalcea. 2018.
\newblock Factors influencing the surprising instability of word embeddings.
\newblock In \emph{NAACL-HLT}.

\bibitem[{Dai et~al.(2019)Dai, Yang, Yang, Carbonell, Le, and
  Salakhutdinov}]{Dai2019TransformerXLAL}
Zihang Dai, Zhilin Yang, Yiming Yang, Jaime~G. Carbonell, Quoc~V. Le, and
  Ruslan Salakhutdinov. 2019.
\newblock Transformer-xl: Attentive language models beyond a fixed-length
  context.
\newblock In \emph{ACL}.

\bibitem[{Grave et~al.(2016)Grave, Joulin, Cisse, Grangier, and
  Jegou}]{Grave2016EfficientSA}
Edouard Grave, Armand Joulin, Moustapha Cisse, David Grangier, and Herve Jegou.
  2016.
\newblock Efficient softmax approximation for gpus.
\newblock \emph{ArXiv}, abs/1609.04309.

\bibitem[{Merity et~al.(2017)Merity, Keskar, and
  Socher}]{Merity2017RegularizingAO}
Stephen Merity, Nitish~Shirish Keskar, and Richard Socher. 2017.
\newblock Regularizing and optimizing lstm language models.
\newblock \emph{ArXiv}, abs/1708.02182.

\bibitem[{Merity et~al.(2016)Merity, Xiong, Bradbury, and
  Socher}]{Merity2016PointerSM}
Stephen Merity, Caiming Xiong, James Bradbury, and Richard Socher. 2016.
\newblock Pointer sentinel mixture models.
\newblock \emph{ArXiv}, abs/1609.07843.

\bibitem[{Mikolov et~al.(2010)Mikolov, Karafi{\'a}t, Burget, Cernock{\'y}, and
  Khudanpur}]{Mikolov2010RecurrentNN}
Tomas Mikolov, Martin Karafi{\'a}t, Luk{\'a}s Burget, Jan Cernock{\'y}, and
  Sanjeev Khudanpur. 2010.
\newblock Recurrent neural network based language model.
\newblock In \emph{INTERSPEECH}.

\bibitem[{Mimno and Thompson(2017)}]{Mimno2017TheSG}
David~M. Mimno and Laure Thompson. 2017.
\newblock The strange geometry of skip-gram with negative sampling.
\newblock In \emph{EMNLP}.

\bibitem[{Radford et~al.(2019)Radford, Wu, Child, Luan, Amodei, and
  Sutskever}]{Radford2019LanguageMA}
Alec Radford, Jeffrey Wu, Rewon Child, David Luan, Dario Amodei, and Ilya
  Sutskever. 2019.
\newblock Language models are unsupervised multitask learners.

\bibitem[{Stolcke(2002)}]{Stolcke2002SRILMA}
Andreas Stolcke. 2002.
\newblock Srilm - an extensible language modeling toolkit.
\newblock In \emph{INTERSPEECH}.

\bibitem[{Yang et~al.(2017)Yang, Dai, Salakhutdinov, and
  Cohen}]{Yang2017BreakingTS}
Zhilin Yang, Zihang Dai, Ruslan Salakhutdinov, and William~W. Cohen. 2017.
\newblock Breaking the softmax bottleneck: A high-rank rnn language model.
\newblock \emph{ArXiv}, abs/1711.03953.

\bibitem[{Zaremba et~al.(2014)Zaremba, Sutskever, and
  Vinyals}]{Zaremba2014RecurrentNN}
Wojciech Zaremba, Ilya Sutskever, and Oriol Vinyals. 2014.
\newblock Recurrent neural network regularization.
\newblock \emph{ArXiv}, abs/1409.2329.

\end{thebibliography}
\bibliographystyle{acl_natbib}

 \textcolor{white} {x} \\
 \textcolor{white} {x} \\
 \textcolor{white} {x} \\
 \textcolor{white} {x} \\
 \textcolor{white} {x} \\
 \textcolor{white} {x} \\
 \textcolor{white} {x} \\
 \textcolor{white} {x} \\
 \textcolor{white} {x} \\
 \textcolor{white} {x} \\
 \textcolor{white} {x} \\
 \textcolor{white} {x} \\
 \textcolor{white} {x} \\
 \textcolor{white} {x} \\
 \textcolor{white} {x} \\
 \textcolor{white} {x} \\
 \textcolor{white} {x} \\
 \textcolor{white} {x} \\
 \textcolor{white} {x} \\
 \textcolor{white} {x} \\
 \textcolor{white} {x} \\
 \textcolor{white} {x} \\
 \textcolor{white} {x} \\
 \textcolor{white} {x} \\
 \textcolor{white} {x} \\
 \textcolor{white} {x} \\
 \textcolor{white} {x} \\
 \textcolor{white} {x} \\
 \textcolor{white} {x} \\
 \textcolor{white} {x} \\
 \textcolor{white} {x} \\
 
\vspace{5in}

\newpage

\appendix

\section{Proof of Theorems}
\label{sec:appendix}

\textbf{Proof of Theorem 1.}

Let $P = \{x_1, \dots , x_N \}$ be the set of all words. We can form the convex hull of this set. If $p$ is interior, then for all $v$, there exists an $x_i \in P$ such that $\inner{v}{x_i - p} > 0$. We argue by contradiction. Suppose that $p$ is interior and that for all $v$, we have that $\inner{v}{x_i - p} \le 0$ for all $x_i \in P$. This implies that all points in our set $P$ lay strictly on one side of the hyperplane made perpendicular to $v$ through $p$. This would imply that $p$ was actually on the convex hull, a contradiction.

This implies that for any test point $h$, an interior point will be bounded by at least one point in $P$. That is $\inner{h}{ p} < \inner{h}{x_i}$ for some $x_i \in P$. Plugging into the softmax function we see that:

\begin{align}
\mathbb{P} (p) & = \frac{\exp(\inner{h}{p})}{\exp(\inner{h}{p}) + \sum_{j \neq p} \exp( \inner{h}{x_j}) } \nonumber \\ & \le \frac{1} {1 +  \exp( \inner{h}{x_i-p}) } \nonumber \\
\nonumber \end{align}

Letting $\| h\| \to \infty$ shows that $\mathbb{P}(p) \to 0$. This shows that interior points are probability deficient. We also note that letting $\| h \| \to 0$ gives the base probability $\mathbb{P}(p) = 1/|P|$.

The contrapositive of the above statement implies that if $ \nexists \ v$, where $\forall \ x_i \in p$ we have $\inner{v}{x_i - p} \le 0$, then $p$ is on the convex hull. In fact, we also note that if $p$ was a vertex, the inequality would be strict, which implies that one can find a test point such that the probability $\mathbb{P}(p) \to 1$.

The more interesting case is if the point $p$ is on the convex hull, but not a vertex. In this case we define the set $\Omega(p,h) = \{x_i \in P \ | \ \inner{h}{p - x_i} = 0 \}$. This corresponds to the set of points lying directly on the hyperplane perpendicular to $h$, running through $p$. This set is nonempty. Then we see that:

\begin{align}
\mathbb{P} (p) & = \frac{\exp(\inner{h}{p})}{ \sum_j \exp( \inner{h}{x_j}) } \nonumber \\
& \le  \frac{\exp(\inner{h}{p})}{ \sum_{j \in \Omega(p,h)} \exp( \inner{h}{x_j}) } \nonumber \\
&= \frac{1}{|\Omega(p,h)|} \nonumber \\
\nonumber \end{align}

\textcolor{white} {x} \\
 \section{3D Numerical Illustration}

\begin{figure}[h!]
\includegraphics[scale=0.50]{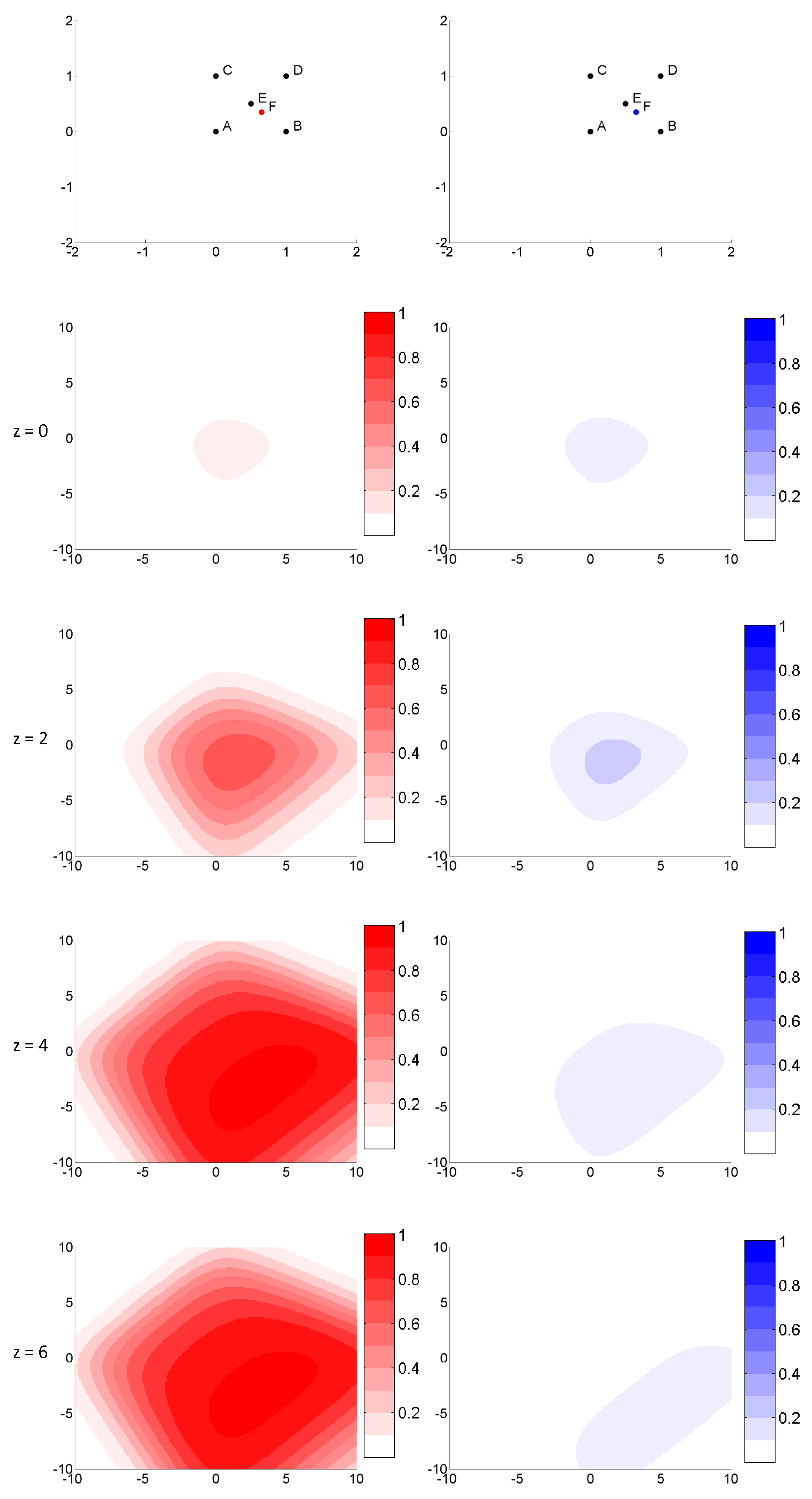}
\centering
\caption{\textbf{Numerical Illustration in 3D} {The top panels show six words in a flattened cross-section of 3D space.  Points $A$, $B$, $C$, $D$ and $E$ are embedded at $(0,0,0)$, $(1,0,0)$, $(0,1,0)$, $(1,1,0)$ and $(0.5,0.5,1)$ respectively.  In the top-left panel, $F$ is embedded outside of the convex hull at $(0.65,0.35,1.5)$, and in the top-right panel $F$ is embedded inside of the convex hull at $(0.65,0.35,0.5)$.  Subsequent panels show cross sections of the probability of $F$ for test points in the plains $z=\{0.0, 2.0, 4.0, 6.0\}$, numerically illustrating the stolen probability effect in 3D.}}
\label{fig:f_apex_b}
\end{figure}

\end{document}